\documentclass{llncs}

\usepackage[utf8]{inputenc}
\usepackage[T1]{fontenc}
\usepackage[font=scriptsize, farskip=5pt]{subfig}
\usepackage{graphicx}
\usepackage{url}
\usepackage{xcolor}
\usepackage{wrapfig}
\usepackage{algorithmicx}
\usepackage{tabularx}
	\newcolumntype{C}[1]{>{\centering\arraybackslash}p{#1}}
	\newcolumntype{R}[1]{>{\raggedleft\arraybackslash}p{#1}}
	\newcolumntype{L}[1]{>{\raggedright\arraybackslash}p{#1}}
\usepackage{booktabs}
\usepackage{amssymb}
\usepackage{amsmath}
\usepackage{siunitx}
\DeclareSIUnit{\calorie}{cal}
\DeclareSIUnit{\Calorie}{\kilo\calorie}

\usepackage{microtype}
\usepackage{xparse}
\usepackage{xspace}

\usepackage{my-tkz-kiviat}
\usepackage{pgfplotstable}
\pgfplotsset{compat=1.14}
\usetikzlibrary{arrows, positioning}

\definecolor{mylinkcolor}{rgb}{0,0,.5} %
\definecolor{mycitecolor}{rgb}{0,0.5,0} %
\usepackage[
colorlinks=true,
breaklinks=true,
linkcolor=mylinkcolor,
menucolor=mylinkcolor,
urlcolor=mylinkcolor,
citecolor=mycitecolor, %
]{hyperref}
\usepackage[capitalize]{cleveref} %

\newcommand{\R}{\mbox{$\mathbb{R}$}}

\newcommand{\etal}{\emph{et al.}\xspace}

\newcommand{\Mpro}{M\textsuperscript{pro}\xspace}

\newcommand{\capsacronym}[1]{{\footnotesize \mbox{#1}}\xspace}
\newcommand{\SMILES}{\capsacronym{SMILIES}}
\newcommand{\SELFIES}{\capsacronym{SELFIES}}
\newcommand{\MOSES}{\capsacronym{MOSES}}
\newcommand{\sarscov}{\capsacronym{SARS-CoV}}
\newcommand{\sarscovtwo}{\capsacronym{SARS-CoV-2}}
\newcommand{\covid}{\capsacronym{COVID-19}}
\newcommand{\EMOA}{\capsacronym{EMOA}}

\def\currenttsvfile{}
\NewDocumentCommand{\drawKdiagram}{ O{} m }{
	\def\currenttsvfile{data/#2}
	\tkzKiviatDiagramFromFile[
		label distance=5mm,
		gap=1,
		space=.8,
		label space=2,
		lattice=10,
		#1
	]{\currenttsvfile}
}
\NewDocumentCommand{\drawKline}{ m m O{} }{
	\tkzKiviatLineFromFile[
		semithick,
		color      = #2,
		mark       = none,
		fill       = #2!20,
		#3
	]{\currenttsvfile}{#1}
}
\NewDocumentCommand{\drawKlines}{ m }{
	\ifthenelse{#1 > 1}{\drawKline{1}{blue}
	\ifthenelse{#1 > 2}{\drawKline{2}{red}
	\ifthenelse{#1 > 3}{\drawKline{3}{yellow}
	\ifthenelse{#1 > 4}{\drawKline{4}{green}
	\ifthenelse{#1 > 5}{\drawKline{5}{orange}
	\ifthenelse{#1 > 6}{\drawKline{6}{brown}
	\ifthenelse{#1 > 7}{\drawKline{7}{cyan}
	\ifthenelse{#1 > 8}{\drawKline{8}{violet}
	\ifthenelse{#1 > 9}{\drawKline{9}{black}
	\ifthenelse{#1 > 10}{\drawKline{10}{gray}
	
	\ifthenelse{#1 > 11}{\drawKline{11}{blue}[dashed]
	\ifthenelse{#1 > 12}{\drawKline{12}{red}[dashed]
	\ifthenelse{#1 > 13}{\drawKline{13}{yellow}[dashed]
	\ifthenelse{#1 > 14}{\drawKline{14}{green}[dashed]
	\ifthenelse{#1 > 15}{\drawKline{15}{orange}[dashed]
	\ifthenelse{#1 > 16}{\drawKline{16}{brown}[dashed]
	\ifthenelse{#1 > 17}{\drawKline{17}{cyan}[dashed]
	\ifthenelse{#1 > 18}{\drawKline{18}{violet}[dashed]
	\ifthenelse{#1 > 19}{\drawKline{19}{black}[dashed]
	\ifthenelse{#1 > 20}{\drawKline{20}{gray}[dashed]
	}{}}{}}{}}{}}{}}{}}{}}{}}{}}{}}{}}{}}{}}{}}{}}{}}{}}{}}{}}{}
}

\NewDocumentCommand{\kiviatAndFormula}{ m m m m }{
	\subfloat[#1]{%
	    \begin{tikzpicture}
 			\drawKdiagram[scale=0.09, label space=3, label style={font=\scriptsize}]{champions/#2.dat}
 			\drawKline{1}{blue}
	    	\node[minimum width=.2\textwidth, minimum height=0.22\textwidth] (a) at (300mm, 0) {\includegraphics[width=.18\textwidth, height=.2\textwidth, keepaspectratio]{pics/champions/#2.pdf}};
	    	\node[anchor=north east, text width=.48\textwidth, align=center, inner sep=0pt, font=\scriptsize] (b) at ([yshift=-10mm] a.south east) {#3};
	    \end{tikzpicture}%
	}
}

\begin{document}

\date{}

\title{Evolutionary Multi-Objective Design of SARS-CoV-2 Protease Inhibitor Candidates}

\author{Tim Cofala$^1$, Lars Elend$^1$, Philip Mirbach$^1$, Jonas Prellberg$^1$, Thomas Teusch$^2$, Oliver Kramer$^1$}

\institute{$^1$Computational Intelligence Lab, Department of Computer Science\\
$^2$Theoretical Chemistry Group, Department of Chemistry\\
University of Oldenburg, Germany\\
\email{<firstname>.<lastname>@uni-oldenburg.de}}
\maketitle
\thispagestyle{empty}

\begin{abstract}
    Computational drug design based on artificial intelligence is an emerging research area.
    At the time of writing this paper, the world suffers from an outbreak of the coronavirus \sarscovtwo.
    A promising way to stop the virus replication is via protease inhibition.
    We propose an evolutionary multi-objective algorithm (EMOA) to design potential protease inhibitors for \sarscovtwo's main protease.
    Based on the \SELFIES representation the EMOA maximizes the binding of candidate ligands to the protein using the docking tool QuickVina 2, while at the same time taking into account further objectives like drug-likeliness or the fulfillment of filter constraints.
    The experimental part analyzes the evolutionary process and discusses the inhibitor candidates.    
	
	\keywords{Evolutionary Multi-objective Optimization \and Computational Drug Design \and SARS-CoV-2}
\end{abstract}

\section{Introduction}
\label{sec:intro}

At the time of writing this paper, researchers around the globe are searching for a vaccine or an effective treatment against the 2019 novel coronavirus (\sarscovtwo).
One strategy to limit virus replication is protease inhibition.
A biomolecule called ligand binds to a virus protease enzyme and inhibits its functional properties.
For \sarscovtwo the crystal structure of its main protease \Mpro has been solved, e.g. by Jin \etal \cite{Jin2020Structure}.
The search for a valid protease inhibitor can be expressed as optimization problem.
As not only the binding of the ligand is an important objective, but also further properties like drug-likeliness or filter properties, we comprise the molecule search problem as multi-objective optimization problem, which we aim to solve with evolutionary algorithms.

This paper is structured as follows.
In \cref{sec:protease} we shortly repeat the basics of protease inhibition and the connection to the novel coronavirus.
\Cref{sec:related} gives an overview of related work on evolutionary molecule design.
In \cref{sec:metrics} we introduce molecule metrics, which we aim to optimize with the \EMOA that is presented in \cref{sec:emoa}.
The experimental part in \cref{sec:exp} presents our experimental results and discusses the evolved molecules.
Conclusions are drawn in \cref{sec:cons}, where also prospective future research directions are presented.

\section{Virus Protease Inhibition}
\label{sec:protease}

\begin{figure}[htb]
    \centering
    \includegraphics[scale=0.33]{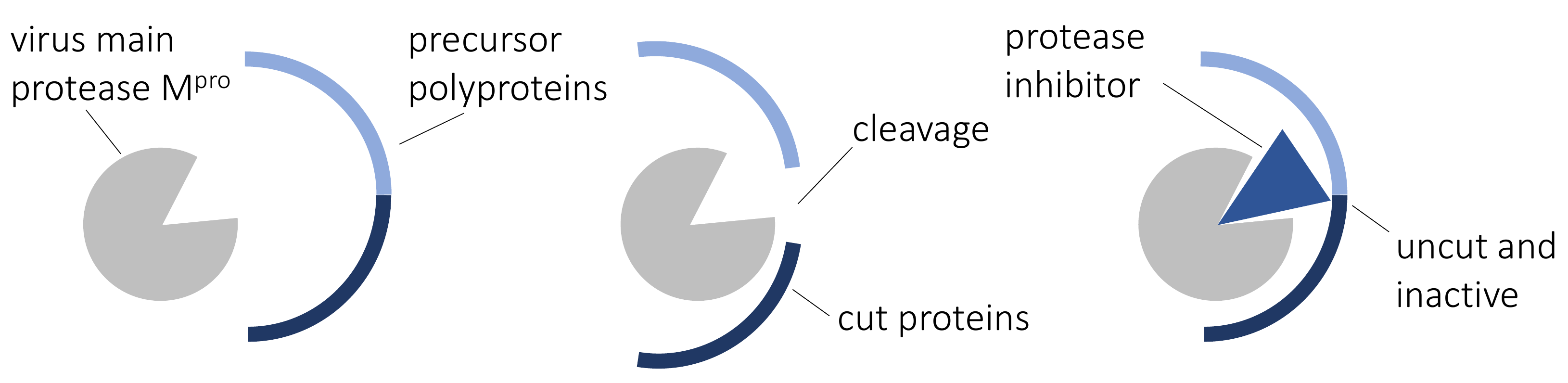}
    \caption{Illustration of (left) protease enzyme with uncut precursor polyproteins, (middle) the cleavage progress, and (right) protease inhibition preventing the cleavage.}
    \label{fig:cleavage}
\end{figure}

As of late 2019, a novel respiratory disease named \covid spread worldwide.
\covid is caused by \sarscovtwo, which belongs to the coronavirus family like the well-known severe acute respiratory syndrome coronavirus (\sarscov).
As RNA virus \sarscovtwo's replication mechanism hijacks the cell mechanisms for replication.
An essential part of the virus replication process is a cleavage process, in which the virus protease enzyme cuts long precursor polyproteins into mature non-structural proteins, see \cref{fig:cleavage}.
If a ligand biomolecule binds to the protease it can prevent and inhibit this cleavage process.
A ligand binds to the target protein in a so-called pocket based on various non-covalent interactions like hydrophobic interactions, hydrogen bonding, $\pi$-stacking, salt bridges, and amide stacking \cite{Freitas2017Systematic}.
With the proper ligand, the protease cleavage process is inhibited, in practice measured by the half maximal inhibitory concentration IC$_{50}$ corresponding to the inhibitory substance quantity needed to inhibit $50\%$ of the protease process.
The protease inhibitor is the target of the drug design process, which we aim to find with evolutionary search.

Computational modeling of protein-ligand binding is a complex process depending on protein-ligand geometry, chemical interactions as well as various constraints and properties like hydration and quantum effects.
Complex molecular dynamics computations are often too expensive in computational drug design.
Instead, docking tools like AutoDock \cite{Morris2009AutoDock4}, see \cref{sec:metrics}, are supposed to be sufficient for a coarse binding affinity estimation based on a simplification of the physical reality.

For \sarscovtwo the crystal structure of its main protease \Mpro is known, e.g. \cite{Dai2020Structurebased,Jin2020Structure,Zhang2020Crystal}.
Various attempts to design inhibitors have been made recently, e.g., based on known protease inhibitors for other viruses \cite{Caly2020FDAapproved,Khaerunnisa2020Potential}, based on virtual screening \cite{Fischer2020Inhibitors}, and computational drug design \cite{Macchiagodena2020Inhibition}.

\section{Related Work}
\label{sec:related}

Methods for de novo drug design can be categorized in different ways \cite{Devi2015Evolutionary,Brown2019GuacaMol}.
Some works construct molecules directly from atoms \cite{Douguet2000Genetic,Nigam2020Augmenting}, while others use chemical fragments as their smallest building block \cite{Pegg2001Genetic}.
The goal also varies among publications and is sometimes to find drugs that bind to a specific protein binding site like in our work or \cite{Pegg2001Genetic,Yuan2020LigBuilder}, and other times the goal is to generate any drug-like molecules as in \cite{Douguet2000Genetic,Polykovskiy2019Molecular}.

ADAPT \cite{Pegg2001Genetic} is a fragment-based method that optimizes for molecule to bind to a specific binding site using a genetic algorithm on an acyclic graph-representation consisting of chemical fragments.
The fitness of a resulting compound is evaluated through a docking simulation with a target protein binding site and common drug-likeness indicators.

On the other hand, Douguet \etal \cite{Douguet2000Genetic} use the \SMILES representation and as such work on the level of atoms instead of fragments.
In contrast to our work their genetic algorithm optimizes for drug-likeness only instead of binding to a specific ligand.
Furthermore, the algorithm is single-objective and simply weighs the different properties in a fitness function using constant coefficients.

Similarly, Nigam \etal \cite{Nigam2020Augmenting} present a genetic algorithm on the \SMILES representation for general molecule design.
The method increases diversity by using a deep neural network as an adaptive fitness function to penalize long-surviving molecules.
In contrast to methods like ours that try to stay inside the distribution of drug-like molecules, the genetic algorithm is free to explore the chemical space in its entirety.

Finally, LigBuilder \cite{Yuan2020LigBuilder} is a software tool for drug design that is based on a genetic algorithm.
It allows optimizing for the interesting quality of binding to multiple targets, which enables tackling more complex diseases with a single drug without the risk of drug-drug interactions that comes with combination drugs (treatment with multiple compounds).

Brown \etal \cite{Brown2004GraphBased} have utilized an approach for multi-objective optimization of molecules applying a graph-based representation of molecules.
The multi-objective evolutionary algorithm applies a Pareto ranking scheme for the optimization process.

Wagner \etal \cite{Wager2016Central} have developed a tool which identifies potential CNS drugs by means of a multi-objective optimization.
The molecules have been optimized for six physical properties.
In contrast to the approach presented here, this tool is not based on evolutionary algorithms but on medical knowledge.

A short review, which focuses on the multi-objective optimization of drugs, is given in \cite{Nicolaou2013Multiobjective}.
In this context, different problem definitions and various Multi-objective optimization methods are summarized.

Since there are many competing approaches to drug and molecule design, benchmarking platforms like \MOSES \cite{Polykovskiy2019Molecular} or GuacaMol \cite{Brown2019GuacaMol} have emerged.
These propose fixed datasets and metrics to measure and compare the generative abilities of different algorithms.
Among the benchmarked algorithms are random sampling, variational autoencoders, generative adversarial networks, Monte-Carlo tree search, and others.

\section{Molecule Design Metrics}
\label{sec:metrics}

In computational drug design, molecule metrics define the optimization objectives.
This section introduces the five metrics our optimization approach is based on.

\subsection{Binding Affinity Scores}

The major objective in protease inhibitor search is the protein-ligand binding affinity.
A widespread tool for this metric is the automated docking tool AutoDock \cite{Morris2009AutoDock4}, which will also be used by the OpenPandamics\footnote{\url{https://www.ibm.org/OpenPandemics}} activities to fight \covid.
AutoDock performs very fast calculations of the binding energy by using grid-based look-up tables.
For this purpose, the protein is embedded in a grid.
The binding energy of all individual atoms of the ligand is calculated at all positions of the grid using semi-empirical force field methods.
Using a Lamarckian genetic algorithm, the best binding position and binding energy of the complete ligand can be determined with the help of the look-up tables.

Through various improvements, the accuracy and especially the performance of AutoDock has been significantly improved.
In AutoDock Vina \cite{Trott2010AutoDock} a hybrid scoring function based on empirical and knowledge-based data is used instead of the force field method.
QuickVina \cite{Trott2010AutoDock} and QuickVina 2 \cite{Alhossary2015Fast} mainly improve the search algorithm by performing the most complex part of the optimization only for very promising ligand positions.
We use QuickVina 2 for the calculation of the binding energies of our proposed ligands, as it provides very good results at high performance.
For the sake of simplicity we will use binding affinity score and docking score synonymously.

The informative value of QuickVina 2 binding scores may be limited due to a simplification of various physical properties, such as the neglect of water molecules and the changing electrical properties of ligand and protein when they interact with each other.
However, it has been shown by Gaillard \cite{Gaillard2018Evaluation} that AutoDock Vina binding scores outperform various computational docking methods and Quickvina 2 achieves very comparable results with Autodock Vina \cite{Alhossary2015Fast}.

\subsection{Quantitative Estimate of Drug-likeness (QED)}

To estimate whether a molecule can be used as a drug, its similarity to other existing drugs can be considered.
This is based on the fact that many important physiochemical properties of drugs follow a certain distribution.
Lipinski's rule of five \cite{Lipinski1997Experimental} which specifies ranges of values for different molecular properties such as size, is frequently used.
A major disadvantage, however, is that this rule is only a rule of thumb and only checks whether its criteria are met or not.
Among modern drugs there are molecules that violate more than one of Lipinski's rules.
A modern approach by Bickerton \etal \cite{Bickerton2012Quantifying} is based on multi-criteria optimization and the principle of desirability.
Instead of a fixed value range, all relevant molecular properties are evaluated by an individual desirability function.
A single score (QED) is then determined by geometrically averaging all desirability functions.

\subsection{Natural Product-likeness (NP)}

In addition to the similarity to known drugs, the similarity to naturally occurring biomolecules (natural products) is also an important metric.
Natural products have numerous bioactive structures that were created and validated by nature in an evolutionary process.
Ertl \etal \cite{Ertl2008Natural} have studied the key differentiating features of natural and synthetic molecules and developed a measure of similarity to natural products.
This score is based on structural characteristics of the molecules, such as the number of aromatic rings and the distribution of nitrogen and oxygen atoms.

\subsection{Medical Chemical Filters}

Medical chemical filters are used to exclude molecules that are toxic due to their structural nature.
Potentially unstable molecules whose metabolites may be toxic are also not suitable as drugs.
We use the MCFs and PAINS filters described by Polykovskiy \cite{Polykovskiy2019Molecular} as a Boolean indicator metric.

\subsection{Synthetic Accessibility (SA)}

For drug design it is not only important to find a molecule with the desired properties, but also a synthesizable one.
Ertl and Schuffenhauer \cite{Ertl2009Estimation} create a method to estimate the synthetic accessibility of drug-like molecules and achieve a high agreement with manual estimations by experts.
Such a method can easily be incorporated into a search process and we use it as one of our optimization goals, too.
A different approach to the synthesis problem is taken by Segler \etal \cite{Segler2018Planning}.
Instead of estimating synthetic accessibility, their symbolic AI driven approach searches for actual synthesis routes of desired target molecules with a combination of Monte Carlo tree search and neural networks encoding rules for reaction centers.

\subsection{Value Ranges} %
\label{sec: metrics value ranges}

\Cref{tab:metric_ranges} shows the value ranges and the optima of the five used metrics.
For our experiments we unify these values to a range of [0, 1], where 0 is the optimum, as we will describe under \cref{sec:fitness evaluation}.

\begin{table}[htb]
    \caption{Value ranges and optimum for used metrics}
    \label{tab:metric_ranges}
    
    \centering
    \small
    \setlength{\tabcolsep}{6pt}
    \begin{tabular}{rccccc}
        \toprule
            & docking score {[}kcal/mol{]} & SA        & QED     & NP        & filters \\
        \midrule 
        value range & $\R$           & $[1, 10]$ & $[0,1]$ & $[-5, 5]$ & $\{0,1\}$ \\
        optimum     & $-\infty$      & 1         & 1       & 5         & 1       \\
    \bottomrule
    \end{tabular}
\end{table}

\section{Evolutionary Molecule Search}
\label{sec:emoa}

This section presents the evolutionary approach for the protease inhibitor design.
For searching in the design space of biomolecules we use evolutionary algorithms (EAs), which are biologically inspired population-based search heuristics.
We employ the evolution strategy oriented $(\mu + \lambda)$ population model \cite{Beyer2002Evolution}.

\begin{figure}[htb]
    \centering
    \begin{tikzpicture}
        \node[above right] (img) at (0,0) {\includegraphics[width=0.13\textwidth]{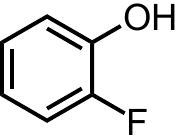}};
        \node [right = of img, yshift=3mm] {\SMILES: Oc1ccccc1F};
        \node [right = of img, yshift=-3mm] {\SELFIES: [O][c][c][c][c][c][c][Ring1][Branch1\_1][F]};
    \end{tikzpicture}
    
    \caption{Molecular structure formula, \SMILES, and \SELFIES of 2-fluorophenol.}
    \label{fig:example_selfies}
\end{figure}

A solution is defined by a string based on the self-referencing embedded strings (\SELFIES) representation \cite{Krenn2020SelfReferencing}, which is an advancement of the simplified molecular-input line-entry system (\SMILES) \cite{Weininger1988SMILES} representation.
Figure \ref{fig:example_selfies} pictures an exemplary molecule with its structural formula and the corresponding \SMILES and \SELFIES representations.
Each string consist of symbols, encoding the occurring atoms, bindings, branches and ring sizes. \SELFIES implements a formal grammar, and the interpretation of a symbol depends on derivation rules and state of derivation.
In contrast to \SMILES, \SELFIES strings are always syntactically correct and therefore always yield valid molecules \cite{Krenn2020SelfReferencing}. 

The EA's initial population consists of individuals with randomly generated strings representation of a fixed length.
Since multiple \SELFIES strings can be translated to the same \SMILES string, the resulting \SMILES string is compared to a global list of all previously generated individuals.
Individuals with a representation that already occurred are discarded and a new individual is generated.
This process is repeated until the population consist of unique individuals and also applies for the generation of offspring individuals.

\subsection{Mutation}
\label{sec:mutation}

Since every \SELFIES string corresponds to a valid molecule and every molecule can be expressed in \SELFIES representation, the design space can be explored by applying random mutations to the strings -- more precisely the \SELFIES symbols of which the string is composed.
Offspring solutions are created by choosing a random individual from the parental population.
Each child is mutated with the following mutation operations with defined probabilities:
\begin{description}
\item[Replacement] is applied independently for every symbol with a probability of $p_{r}$. The symbol is replaced by a random \SELFIES symbol.
\item[Insertion] is applied with probability $p_{i}$. A random symbol is inserted at a random position in the individual's representation.
\item[Deletion] is applied with probability $p_{d}$ and deletes a randomly chosen symbol of the individual's representation.
\end{description}
The new symbols are drawn from a set of symbols inspired by \cite{Krenn2020SelfReferencing}.
This set has been extended with benzene as a separate, composed symbol, to increase the likelihood of its occurrence and ease the generation of complex molecules.
Additionally, each symbol is assigned a weighting parameter to adjust the probability with which the is is randomly selected. 
This weighting can be used to increase the likelihood of more common symbols (e.g. [C]) in contrast to more complex ones (e.g. branches and ring structures).

\subsection{Fitness Evaluation}
\label{sec:fitness evaluation}

For the selection operator the fitness $f(\mathbf{x})$ of each solution candidate is evaluated based on the molecule metrics binding affinity score, QED, filters, NP, and SA introduced in \cref{sec:metrics}. 
To increase the comparability, each metric is scaled to the range between 0 (best possible score)  and 1 (worst possible score).
The binding affinities are scaled with regard to the experimentally chosen minimum of \SI[per-mode=symbol]{-15}{\Calorie\per\mole} and maximum of \SI[per-mode=symbol]{1}{\Calorie\per\mole} and clipped to the range between 0 and 1 with soft clipping \cite{Klimek2018Neural}.

For the single-objective baseline experiments each individual is assigned a single composed fitness value. We use a weighted sum fitness of the $n$ introduced metrics:
\begin{equation}
    f(\mathbf{x}) = \sum_{i=1}^n w_i f_i(\mathbf{x})
\end{equation}
with weights $\mathbf{w}=(0.4,0.15,0.15,0.15,0.15)$ with $i$ corresponding to 1: docking, 2: SA, 3: QED, 4: NP, and 5: filters. The choice of weights is based on preliminary experiment with the objective of putting the highest attention on the docking score, while at the same time considering the other properties.

Individuals are evaluated in parallel.
Therefore, the respective \SELFIES strings are converted to the \SMILES representation.
\MOSES is then used for the calculation of QED, NP, and SA as well as for the application of the \capsacronym{PAINS} and \capsacronym{MCF} filters.
The docking score for each compound is determined by QuickVina 2. 
Therefore, RDKit\footnote{\url{https://www.rdkit.org}} and MGLTools\footnote{\url{http://mgltools.scripps.edu}} are used to generate \capsacronym{PDB} and \capsacronym{PDBQT} files for the respective \SMILES representation. The binding energy is calculated in regards to the \covid \Mpro (PDB ID: 6LU7 \cite{LiuX.20206LU7})\footnote{PDB: protein data base, \url{https://www.rcsb.org}} with the search grid being centered around the native ligand position and sized to $22 \times 24 \times 22\ \si{\angstrom}^3$. The exhaustiveness is maintained at its default value of 8.

\subsection{NSGA-II}

The objectives presented in \cref{sec:metrics} may be contradictory.
For example, in preliminary experiments, we discovered that molecules with high AutoDock binding scores suffer from low QED scores.
As the choice of predefined weights for objectives is difficult in advance, a multi-objective approach may be preferable in practice.
In our multi-objective optimization setting in molecule space $\mathcal{M}$ with fitness functions $f_1, \ldots, f_n$ to minimize we seek for a Pareto set
$\{\mathbf{x}^* \mid \nexists \mathbf{x} \in \mathcal{M}: \mathbf{x} \prec \mathbf{x}^* \}$
of non-dominated solutions, where $\mathbf{x} \prec \mathbf{x}^*$ means $\mathbf{x}$ dominates $\mathbf{x}^*$, i.e.,
$\forall i \in \{1, \ldots, n\}: f_i(\mathbf{x}) \leq f_i(\mathbf{x}^*)$,
while $\exists i \in \{1, \ldots, n\}: f_i(\mathbf{x}) < f_i(\mathbf{x}^*)$.
NSGA-II \cite{Deb2002Fast} is known to be able of approximating a Pareto set with a broad distribution of solutions in objective space, i.e., of the Pareto front.
After non-dominated sorting, $\mu$ non-dominated solutions maximizing the crowding distance, which corresponds to the sum of Manhattan distances between the neighboring solutions in objective space. For comparison of different multi-objective runs we also employ the S-metric measuring the dominated hypervolume in objective space with regards to a dominated reference point \cite{Beume2007SMSEMOA}.

\section{Experiments}
\label{sec:exp}

In this section we experimentally analyze the single-objective and the NSGA-II approaches for the protease inhibitor candidate search.
For the experimental analyses, the following settings are applied. A $(10+100)$-EA is used for the single-objective run i.e., in each generation from 10 parents 100 offspring candidate molecules are generated with the mutation operators introduced in \cref{sec:mutation} with mutation probabilities $p_{r}=0.05$, $p_{i}=0.1$, and $p_{d}=0.1$ applying plus selection.
For multi-objective runs the number of parents is increased to 20 to achieve a broader distribution of solutions in objective space. No crossover is applied. 
Individuals are limited to a length of 80 \SELFIES tokens oriented to the setting by Krenn \etal \cite{Krenn2020SelfReferencing}.
All runs are terminated after 200 generations and are repeated 20 times.

\subsection{Metric Development}

\begin{figure}[htb]
    \centering
    \includegraphics[width=\textwidth]{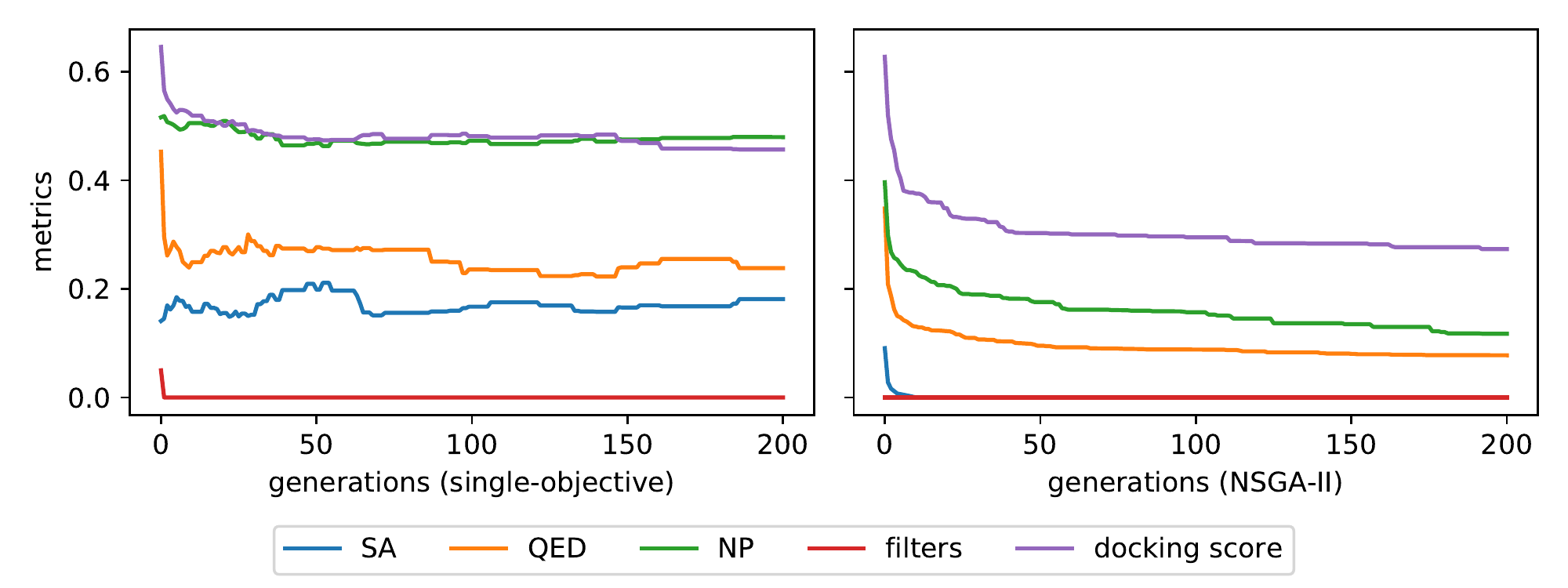}
    \caption{Development of all metrics during (left) single-objective and (right) multi-objective NSGA-II optimization runs.}
    \label{fig:metrics}
\end{figure}

\Cref{fig:metrics} shows the development of the previously explained normalized metrics in single- and multi-objective runs.
For the single-objective runs, the best individuals according to fitness are chosen in each generation and their metrics are averaged over all runs.
The optimization process concentrates on improving docking score, QED, and NP.
As expected, an improvement of one metric may result in a deterioration of another, e.g., as of generation 140, when QED and NP deteriorate in favor of SA and docking score.

For multi-objective runs, the best individuals for each metric are chosen in each generation and then averaged over all runs.
A steady improvement with regard to all objectives is achieved here, but has to be paid with regard to deteriorations in other objectives that are not shown here.

\begin{figure}[htb]
    \centering
    \subfloat[docking score vs. QED]{%
    	\includegraphics[width=0.33\textwidth]{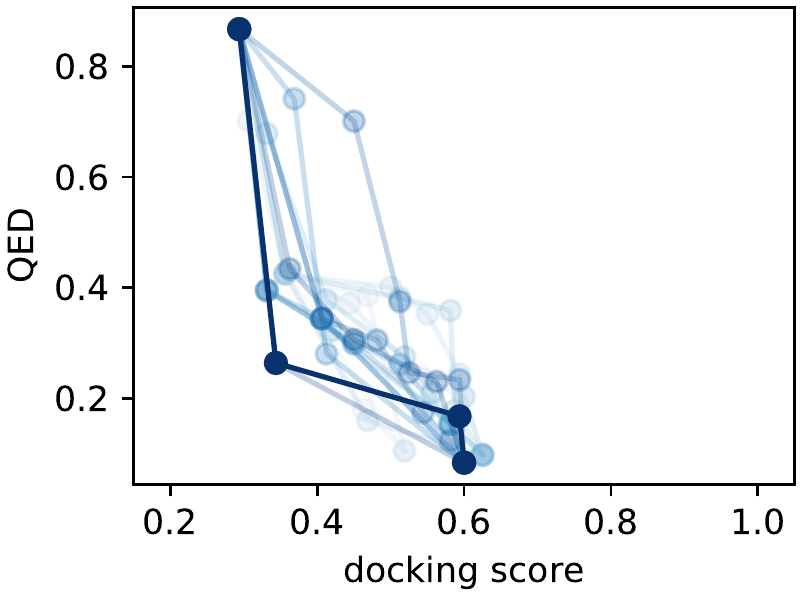}
    }
    \subfloat[docking score vs. NP]{%
    	\includegraphics[width=0.33\textwidth]{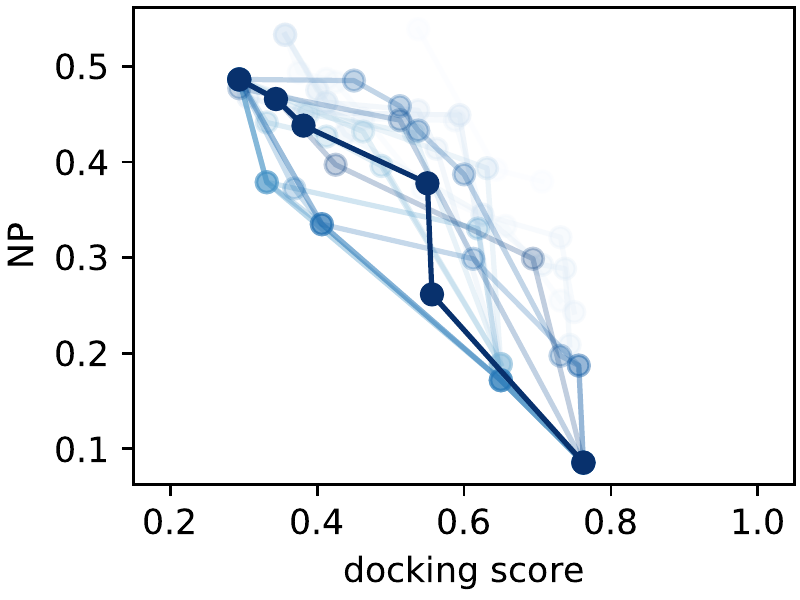}
    }
    \subfloat[docking score vs. SA]{%
        \includegraphics[width=0.33\textwidth]{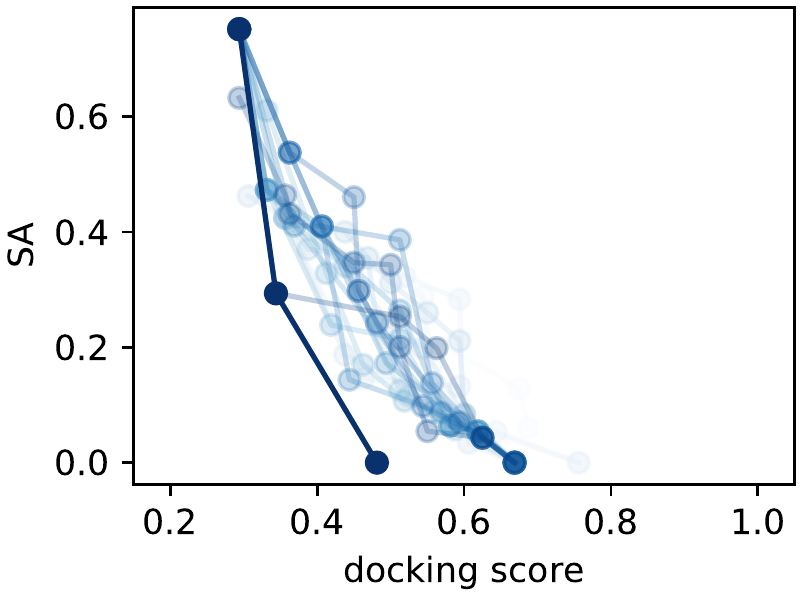}
    }
    \caption{Visualization of typical Pareto fronts evolved with NSGA-II: (a) docking score vs. QED, (b) docking score vs. NP, and (c) docking score vs. SA.}
    \label{fig:pareto}
\end{figure}

\Cref{fig:pareto} shows three different two-dimensional slices of the Pareto front that compare docking score to QED, NP, and SA.
A Pareto front is shown for every 10th generation and their colors start at light blue for the first generation and end at dark blue for the final generation.
The plots illustrate NSGA-II's ability to generate solutions with different degrees of balance between docking score and the plotted metric.

In the course of the optimization process the front of non-dominated solutions has the expected tendency to move towards the lower left.
This is also reflected by the S-metric, which, in average over all runs improves from $0.10 \pm 0.03$ in the first to $0.20 \pm 0.05$ in the last generation.
In the slice plots deteriorations are possible due to improvements in the remaining three objectives.

\begin{table}[htb]
    \caption{Experimental results of weighted-sum single-objective approach, the best values per objective for NSGA-II, the N3 ligand (from PDB 6LU7), and Lopinavir (a prominent drug candidate). Statistical evaluation for the NSGA-II method is calculated based on the best 20 individuals per objective.  $\blacktriangledown$ marks a minimization objective, while $\blacktriangle$ marks a maximization objective.}
    \label{tab:experimental results}

    \includegraphics[width=\textwidth, keepaspectratio]{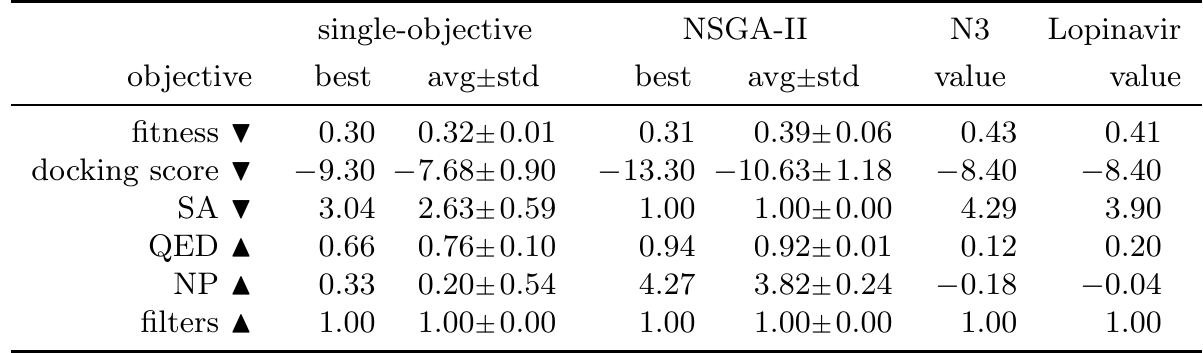}
\end{table}

A comparison of final experimental results of the single-objective and NSGA-II runs is presented in \Cref{tab:experimental results}.
For NSGA-II the best achieved values for each objective are shown corresponding to the corner points of the Pareto front approximation.
For comparison, corresponding metric values are shown for N3 proposed as ligand in the PDB database as well as for Lopinavir, the HIV main protease inhibitor \cite{Kaplan2005Safety}.
Docking scores achieved by the single objective optimization process show that the best values even overcome the scores of N3 and Lopinavir.
Lopinavir and N3 bind similarly strong to \Mpro.
NSGA-II achieves promising values for all metrics. The broad coverage of objective function values offers the practitioner a huge variety of interesting candidates.
However, some of the extreme metric values may sometimes be unpractical, e.g., the outstanding docking score of the best NSGA-II molecule (docking score \SI[per-mode=symbol]{-13.3}{\Calorie\per\mole}) has been achieved by a chemically unrealistic candidate.

\begin{figure}[htb]
    \centering
    \subfloat[single objective]{%
        \resizebox{0.45\textwidth}{!}{%
    	    \begin{tikzpicture}
		        \drawKdiagram[scale=.15]{2020-04-15_19-29-11_0_single_objective_kivat.dat}
		        \drawKlines{10}
	        \end{tikzpicture}%
	    }
    }
    \subfloat[NSGA-II]{%
        \resizebox{0.45\textwidth}{!}{%
    	    \begin{tikzpicture}
		        \drawKdiagram[scale=.15]{2020-04-18_10-33-36_0_nsga2_kiviat.dat}
		        \drawKlines{20}
	        \end{tikzpicture}%
	    }
    }
    
    \caption{%
        Comparison of population of the last generation of typical single-objective (10 molecules) and NGSA-II (20 molecules) runs. Each line represents a molecule candidate.
    }
    \label{fig:spider last generation}
\end{figure}
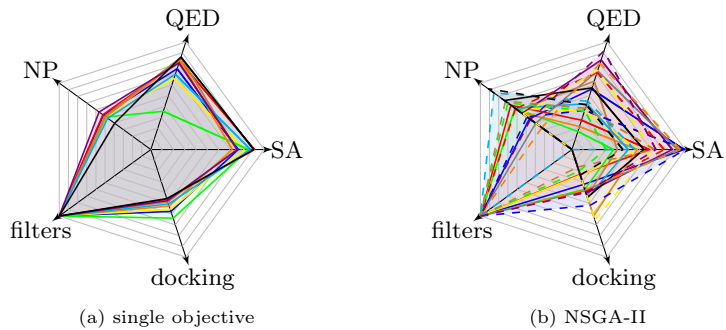

From our observations we conclude that the \SELFIES representation with our mutation operators are able to robustly achieve molecules of a certain quality. However, we expect the quality of the results to improve with mechanisms that allow the development of larger molecules to overcome fitness plateaus and local optima.
Figure \ref{fig:spider last generation} compares the populations of the last generation of a typical single-objective and NSGA-II run. The solutions in the single-objective population are similar to each other, while the solutions in the last NSGA-II population maintain a higher diversity of molecule properties.

\begin{figure}[htb]
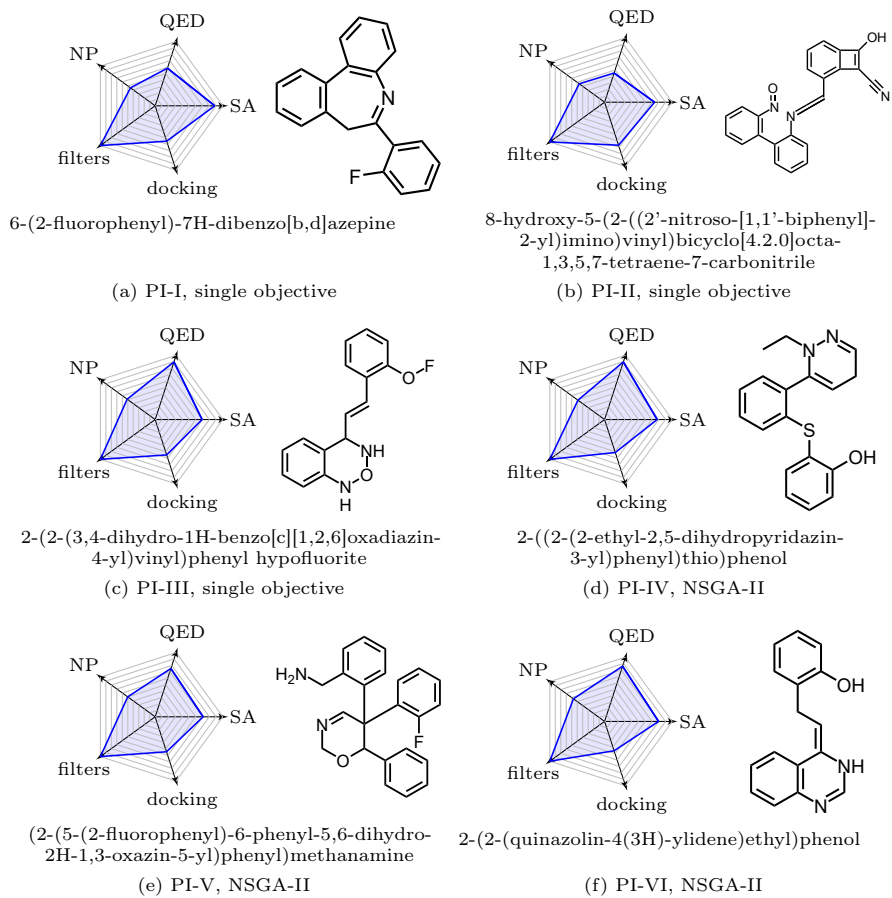

    \kiviatAndFormula{PI-I, single objective}
    {std_indi10546}
    {6-(2-fluorophenyl)-7H-dibenzo[b,d]azepine\newline\newline}
    {Fc1ccccc1C1=Nc2ccccc2-c2ccccc2C1}
    \kiviatAndFormula{PI-II, single objective}
    {std_indi14635}
    {8-hydroxy-5-(2-((2'-nitroso-[1,1'-biphenyl]-2-yl)imino)vinyl)bicyclo[4.2.0]octa-1,3,5,7-tetraene-7-carbonitrile}
    {N\#CC1=C(O)c2cccc(C=C=Nc3ccccc3-c3ccccc3N=O)c21}
    
    \kiviatAndFormula{PI-III, single objective}
    {std_indi18596}
    {2-(2-(3,4-dihydro-1H-benzo[c][1,2,6]oxadiazin-4-yl)vinyl)phenyl hypofluorite}
    {FOc1ccccc1C=CC1NONc2ccccc21}
	\kiviatAndFormula{PI-IV, NSGA-II}
	{nsga2_indi13575}
	{2-((2-(2-ethyl-2,5-dihydropyridazin-3-yl)phenyl)thio)phenol}
	{CCN1N=CCC=C1c1ccccc1Sc1ccccc1O}
    
    \kiviatAndFormula{PI-V, NSGA-II}
    {nsga2_indi18186}
    {(2-(5-(2-fluorophenyl)-6-phenyl-5,6-dihydro-2H-1,3-oxazin-5-yl)phenyl)methanamine}
    {NCc1ccccc1C1(c2ccccc2F)C=NCOC1c1ccccc1}
    \kiviatAndFormula{PI-VI, NSGA-II}
    {nsga2_indi13661}
    {2-(2-(quinazolin-4(3H)-ylidene)ethyl)phenol\newline}
    {Oc1ccccc1CC=C1NC=Nc2ccccc21}
    
    \caption{%
        Exemplary protease inhibitors with properties presented as radar plot, structural formula, and chemical name, a-c: single-objective, d-f: NSGA-II results.
    }
    \label{fig:candidates}
\end{figure}

\subsection{Candidate Comparison}

In the following we present interesting protein inhibitor candidates evolved with the single- and multi-objective approaches. In our experiments we made three main observations.
The molecules generated have a strong tendency to contain aromatic ring structures. Candidates with good drug-likeliness are comparatively short.
Candidates with high docking scores often have unrealistic geometries.

In \Cref{fig:candidates} we present a list of six promising protease inhibitors (PI) candidates with properties as radar plots, structural formulas, and chemical names.
PI-I (a) to PI-III (c) are results from single-objective runs, while PI-IV (c) to PI-VI (f) show candidates generated by NSGA-II.
Points near the border of the radar plot represent better values, e.g., a zero value lies at the corner of a plot.
All candidates fulfill the filter condition.
PI-1 achieves a high SA value with a reasonable docking score.
PI-II achieves an excellent docking score with $-9.7\,$kcal/mol.
PI-III, PI-IV, and PI-VI achieve excellent drug-likeliness QED with good docking results around $-7.0\,$kcal/mol.
An interesting candidate balancing all objectives is PI-V with docking score $-7.7\,$kcal/mol and QED value of $0.75$.

\begin{figure}[htb]
    \centering
    \subfloat[PI-I in \Mpro pocket]{%
    	\includegraphics[width=0.5\textwidth, height=30mm, keepaspectratio]{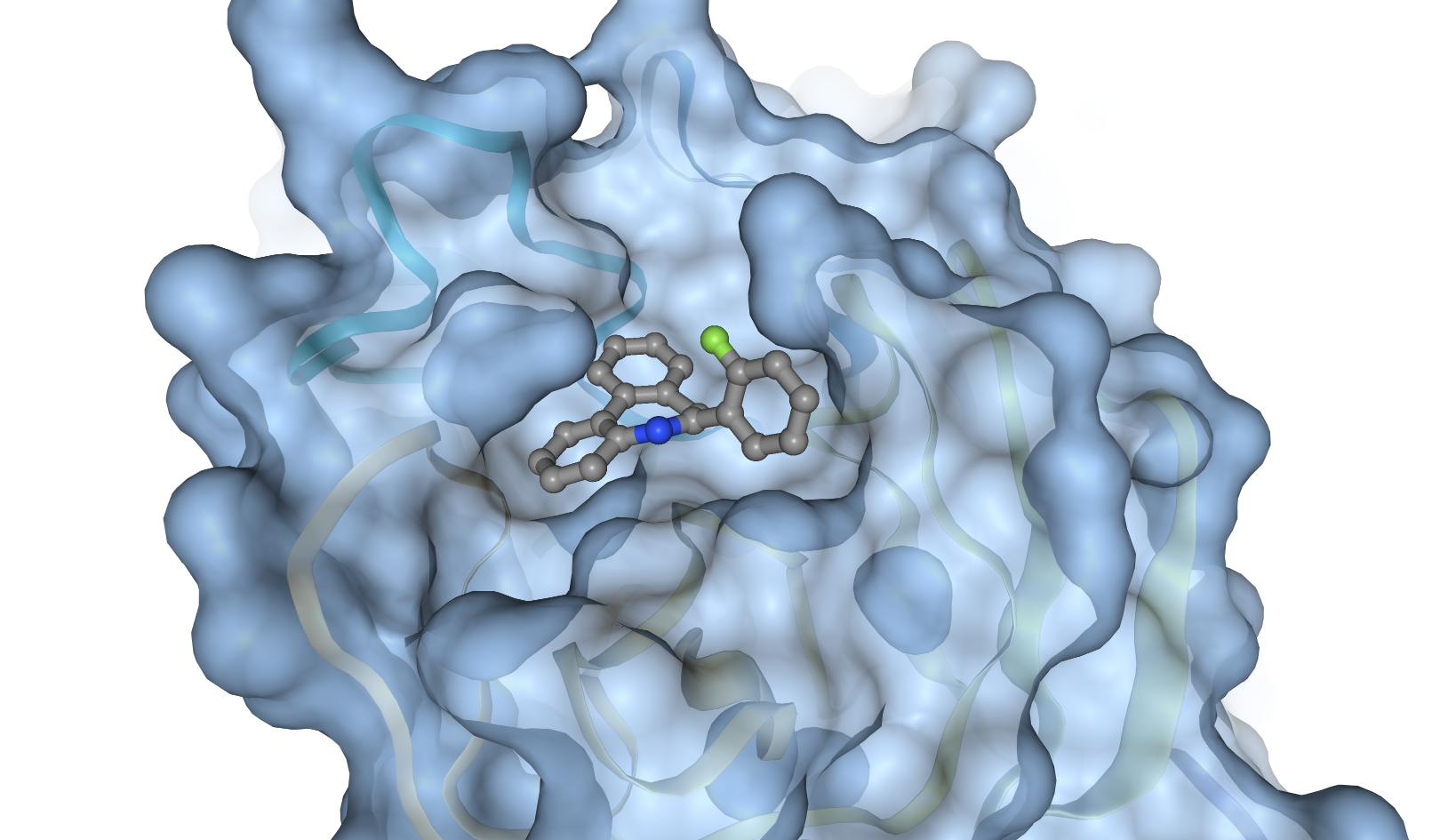}
    }
    \subfloat[PI-V in \Mpro pocket]{%
    	\includegraphics[width=0.5\textwidth, height=30mm, keepaspectratio]{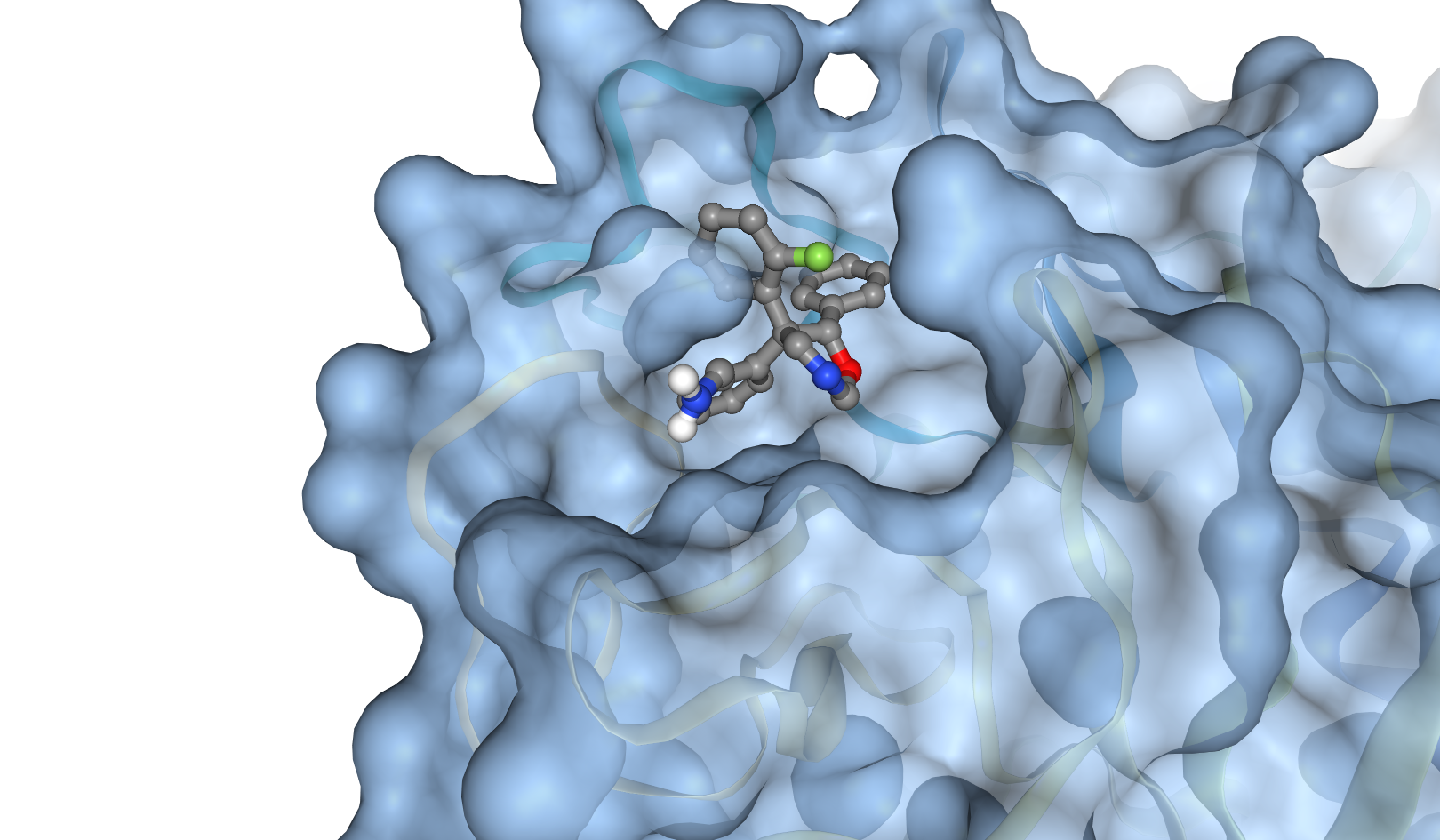}
    }
    \caption{Visualization of PI-I and PI-V docked to the pocket of \sarscovtwo's \Mpro using NGLview \cite{Nguyen2018NGLview}.}
    \label{fig:pocket}
\end{figure}

Last, we visualize how the ligand candidates are located in the \Mpro protein pocket optimized by QuickVina 2.
\Cref{fig:pocket} shows candidates (a) PI-I and (b) PI-V in their \Mpro pockets.

\section{Conclusion}
\label{sec:cons}

In this paper we introduced an evolutionary multi-objective approach to evolve protein inhibitor candidates for the \Mpro of \sarscovtwo, which could be a starting point for drug design attempts, aiming at optimizing the QuickVina 2-based protein-ligand binding scores and further important objectives like QED drug-likeliness and filter properties.
In the experimental part we have shown that the evolutionary processes are able to evolve interesting inhibitor candidates.
Many of them achieve promising metrics with ordinary structures, but also unconventional candidates have been evolved that may be worth for a deeper analysis.
As the informative value of QuickVina 2 binding scores and also the further metrics may be limited in practice, we understand our approach as AI-assisted virtual screening of the chemical biomolecule space.

Future research will focus on the improvement of protein-ligand models for more detailed and more efficient binding affinity models.
Further, we see potential to improve the \SELFIES representation in terms of bloated strings that represent comparatively small molecules and mechanisms to guarantee their validity.

\section*{Acknowledgements}

We would like to thank Ahmad Reza Mehdipour, Max Planck Institute of Biophysics, Frankfurt, Germany, for useful suggestions and comments that helped to improve this manuscript.

\clearpage
\appendix

\bibliographystyle{spmpsci}
\bibliography{literature}

\end{document}